\documentclass[letterpaper, 10 pt, conference]{ieeeconf}

\usepackage{graphicx}
\usepackage{url}
\usepackage{acronym}
\usepackage{stfloats}
\usepackage[tight]{subfigure}
\usepackage{amsmath}
\usepackage{gensymb}
\usepackage{amssymb}
\usepackage{amsfonts}
\usepackage{amsopn}
\usepackage{psfrag}
\usepackage{mdwtab}
\usepackage{enumerate}
\usepackage{mdwlist}
\usepackage{accents}
\usepackage{enumerate}
\usepackage{color}
\usepackage{algorithmicx}

\usepackage{algorithmicx}
\usepackage{algorithm}
\usepackage{epstopdf}
\usepackage{algpseudocode}
\usepackage[english]{babel}
\usepackage[utf8]{inputenc}
\usepackage[normalem]{ulem}
\usepackage[font=footnotesize]{caption}

\DeclareMathOperator*{\minimize}{minimize}

\acrodef{JPS}[\textsc{JPS}]{Jump Point Search}
\acrodef{ROM}[\textsc{ROM}]{Reduced Order Model}
\acrodef{RRT}[\textsc{RRT}]{Rapidly-exploring Random Trees}
\acrodef{LIP}[\textsc{LIP}]{Linear Inverted Pendulum}
\acrodef{COM}[\textsc{COM}]{Center of Mass}
\acrodef{ZMP}[\textsc{ZMP}]{Zero Moment Point}
\acrodef{DCM}[\textsc{DCM}]{Divergent Component of Motion}
\acrodef{CCM}[\textsc{CCM}]{Convergent Component of Motion}
\acrodef{SDP}[\textsc{SDP}]{Semidefinite Program}
\acrodef{CBF}[\textsc{CBF}]{Control Barrier Function}
\acrodef{MPC}[\textsc{MPC}]{Model Predictive Control}
\acrodef{QP}[\textsc{QP}]{Quadratic Program}
\acrodef{SC}[\textsc{SC}]{Safe Corridor}
\acrodef{SWC}[\textsc{SWC}]{\emph{Safe Walking Corridors}}
\acrodef{AASC}[\textsc{AASC}]{\emph{Axis Aligned Safe Corridors}}
\acrodef{IRIS}[\textsc{IRIS}]{Iterative Regional Inflation by Semidefinite Programming}

\newtheorem{remark}{Remark}

\IEEEoverridecommandlockouts

\title{\LARGE \bf
Capturability-Aware Footstep Planning in Cluttered Environments: \\ A Recursively Feasible Sequential MPC on Axis-Aligned Safe Corridors
}

\author{A. Kyriakopoulos$^{1}$, S. S. Kumbhar$^{2}$, K. S. Narkhede$^{2}$, A. M. Kulkarni$^{2}$ and I. Poulakakis$^{3}$
\thanks{$^{1}$A. Kyriakopoulos was with the School of Mechanical Engineering, National Technical University of Athens, Athens, Greece, when this work was conducted. He is now with the Department of Mechanical Engineering, University of California, Berkeley, CA 94720, USA; {\tt\small akyriakopoulos@berkeley.edu}}   
\thanks{$^{2}$S. S. Kumbhar, K. S. Narkhede, and A. M. Kulkarni are with the Department of Mechanical Engineering, University of Delaware, Newark, DE 19716, USA; {\tt\small \{shubhamk, kunalnk, amkulk\}@udel.edu}}%
\thanks{$^{3}$I. Poulakakis is with the School of Mechanical Engineering, National Technical University of Athens, Athens, Greece, and HERON---Hellenic Center of Excellence in Robotics, Athena
Research Center, Athens, Greece;
{\tt\small \{poulakas@mail.ntua.gr, i.poulakakis@athenarc.gr\}}}
}

\begin{document}

\maketitle
\thispagestyle{empty}
\pagestyle{empty}

%%%%%%%%%%%%%%%%%%%%%%%%%%%%%%%%%%%%%%%%%%%%%%%%%%%%%%%%%%%
\begin{abstract}
A fundamental aspect of humanoid footstep planning is to jointly guarantee obstacle avoidance and locomotion stability. This paper presents a capturability-aware sequential Model Predictive Control (MPC) framework for Linear Inverted Pendulum (LIP)-based humanoid footstep planning in the presence of obstacles. The proposed method combines obstacle avoidance with capturability constraints through safe corridors constructed as sequences of intersecting rectangles, yielding a sequential MPC framework with recursive feasibility guarantees for the LIP-based planning layer. The planner autonomously adapts to constrained scenarios, switching to in-place walking when obstructions---such as doors assumed to be open during planning but unexpectedly found to be closed during execution---block forward motion. Performance is largely independent of the prediction horizon, allowing high success rates while keeping the computational cost low. The approach is validated in high-fidelity simulations with the humanoid robot Digit, demonstrating a practical solution for capturable humanoid footstep planning in cluttered environments.
\end{abstract}
%%%%%%%%%%%%%%%%%%%%%%%%%%%%%%%%%%%%%%%%%%%%%%%%%%%%%%%%%%

%=========================================================
\section{Introduction}
\label{sec:introduction}
%=========================================================

Ensuring safe navigation in cluttered, human-centric environments is a necessity for humanoid robots. A key challenge is footstep planning, which must produce feasible step sequences that guide the robot to its goal while ensuring both collision avoidance and reliable locomotion. Existing methods often rely on \ac{ROM} to generate obstacle-free plans without explicit locomotion stability guarantees, whereas approaches focused on the latter typically ignore obstacles. This paper addresses this gap by integrating capturable locomotion with safe corridors, resulting in a capturability-aware sequence of \ac{MPC} schemes for footstep planning with recursive feasibility guarantees.

For navigation in cluttered environments, a common strategy is to decompose the free space into a \emph{\ac{SC}}; a sequence of overlapping convex polytopes; that allows the robot to move from its initial location to the destination without collisions. \ac{SC}s have been widely applied in motion planning for quadrotors~\cite{Deits2015ICRA, Liu2017RAL}, ground robots~\cite{Gao2025TRO}, and humanoids and bipeds~\cite{Narkhede2022, Narkhede2024JDSMC}. They can be constructed from a seed point using \ac{IRIS}~\cite{Deits2015IRIS, Gao2025TRO}, path-based ellipsoidal dilation~\cite{Liu2017RAL}, or voxel-based methods~\cite{Gao2018ICRA, Gao2020TeachRepeat}. The \ac{AASC} in~\cite{Gao2018ICRA} are particularly appealing for their separable box constraints and efficient collision checking, while encompassing a wide range of environments where obstacles can be effectively approximated by rectangular bounds. By design, however, \acp{SC} are purely geometric constraints that ensure collision avoidance, and thus produce motion plans that do not explicitly account for kinodynamic effects.

%fig
\begin{figure}[t]
\vskip +5pt
\centering
{
\includegraphics[width=\columnwidth]{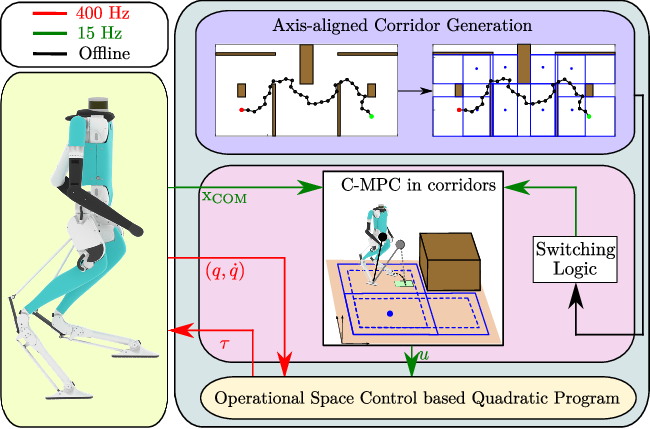}
}
\vskip -15pt
\caption{Overview of the proposed capturability-aware planning framework: Axis-aligned \ac{SC}s enable a sequential \ac{MPC} scheme that reliably drives the robot to its destination, avoiding obstacles and ensuring recursive feasibility.}
\label{fig:concept}
\vskip -15pt
\end{figure}
%fig

Turning our attention to bipedal robots, \ac{MPC} has emerged as a prominent framework for motion planning, as it naturally integrates system dynamics and constraints within a unified optimization problem~\cite{Katayama2023AdvRob}. Combined with reduced-order models such as the \ac{LIP}~\cite{Kajita2001IROS}, \ac{MPC}-based approaches have been applied to address a wide range of challenges, from tracking velocity commands~\cite{Ijspeert2014RSS, Tsagarakis2019IROS} to navigating cluttered environments~\cite{Xiong2021ICRA, Narkhede2022}, and overtaking moving obstacles~\cite{Narkhede2023IROS}. The work of~\cite{Narkhede2022} leveraged the decomposition provided by \ac{SC}s to recast global navigation as a sequence of localized \ac{MPC} subproblems, thereby reducing computational complexity by restricting planning to the robot's immediate neighborhood. Nevertheless, a key challenge in \ac{MPC}-based planners is ensuring recursive feasibility~\cite{borrelli_bemporad_morari_2017, Mayne}, so that feasibility at the current step guarantees feasibility at all future steps.

Recursive feasibility is typically ensured by appropriately designing the terminal components of the \ac{MPC}, either via a terminal constraint or a terminal cost~\cite{borrelli_bemporad_morari_2017, Mayne}. Building on the concept of capturability~\cite{Koolen2011}, \ac{LIP}-based \ac{MPC} approaches have incorporated terminal capture-point constraints to guarantee stability and recursive feasibility~\cite{Sherikov2014Whole, Ciocca2017Strong}, though such constraints can hinder convergence for short prediction horizons~\cite{krause2012Stabilization}. The work in~\cite{Scianca2020} established recursive feasibility and stability for given footsteps by constraining future \ac{ZMP} velocities, conjecturing values beyond the horizon. Subsequent work in~\cite{Smaldone2021} removed the need for a predefined footstep plan via a step-timing adaptation module. A computationally efficient walking pattern generator was proposed in~\cite{Righetti2020TRO} that ensures gait viability by selecting the next step's location and timing. While these studies address walking in unbounded regions, the presence of environmental boundaries introduces additional constraints that may undermine feasibility. This issue was partially addressed in~\cite{Kumbhar2025ICRA}, which introduced a finite-step capture-region constraint, dynamically updated to allow the robot to stop or advance while preserving recursive feasibility. Nevertheless, ensuring feasible MPC footstep plans in the presence of obstacles remains largely open.

To address this gap, this paper proposes a capturability-aware sequential \ac{MPC} framework for humanoid footstep planning amid obstacles. More specifically, a novel \ac{SC} generation framework based on a sequence of intersecting rectangles, each associated with capturability guarantees, together with a switching policy that provides conditions for preserving capturability during transitions between consecutive rectangles along the corridor. This substantially extends existing approaches by combining obstacle avoidance, absent from~\cite{Kumbhar2025ICRA}, with locomotion-stability considerations not provided by the purely geometric strategy of~\cite{Narkhede2022}.
Building on this construction, we then develop a \ac{LIP}-based sequential \ac{MPC} formulation that guarantees recursive feasibility, thereby jointly ensuring collision-free navigation and capturable footstep planning. 
High-fidelity MuJoCo simulations with Digit, including comparisons with~\cite{Narkhede2022} in randomly generated environments with axis-aligned rectangular obstacles, demonstrate the method's intrinsic safety and graceful degradation: when obstructions such as closed doors prevent progress, the robot \emph{autonomously} slows down and switches to in-place walking, unlike approaches that do not explicitly account for capturability. 
The proposed method provides a practical and reliable solution for capturability-aware humanoid footstep planning in cluttered environments.

%===========================================================
\section{LIP-based Capturability \& Restricted Spaces}
\label{sec:problem-statement}
%===========================================================

This section discusses the formulation of a recursively feasible \ac{LIP}-based \ac{MPC}  footstep planner for navigation in obstacle-free rectangular regions.

%fig
\begin{figure*}[t!]
    \vskip +5pt
    \centering
    \includegraphics[width=\textwidth,height=0.2\textheight]{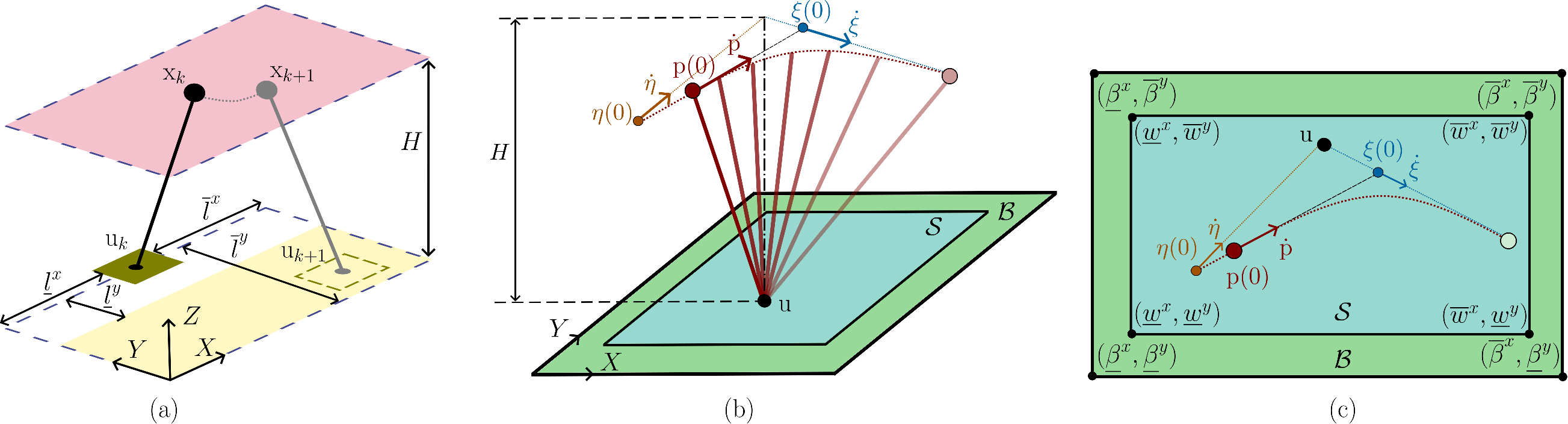}
    % \vspace{-0.2in}
    \caption{(a) Linear Inverted Pendulum (\ac{LIP}) model: \ac{COM} moves in a constant-height plane (pink), with the kinematically feasible foot placement region in yellow. (b) 3D evolution of \ac{COM} (red), \ac{DCM} offset (blue), and \ac{CCM} offset (orange dashed); rectangular regions $\mathcal{B}$ (green) and $\mathcal{S}$ (blue). (c) Top view of \ac{COM}, \ac{DCM}, \ac{CCM} trajectories and the rectangles $\mathcal{B}$, $\mathcal{S}$.
    }
    \label{fig:all-figures}
    \vskip -15pt
\end{figure*}
% fig

%-----------------------------------------------------------
\subsection{The Linear Inverted Pendulum and Capturability}
\label{subsec:lip-capt}
%-----------------------------------------------------------

The Linear Inverted Pendulum (\ac{LIP}) model represents a point mass supported by a massless prismatic leg, with the mass constrained to move on a horizontal plane at constant height $H$; see Fig.~\ref{fig:all-figures}(a). Its state is given by the position $(x,y)\in\mathbb{R}^2$ and velocity $(\dot{x},\dot{y})\in\mathbb{R}^2$ of the mass in the inertial frame, while the stance foot location is given by $(u^x,u^y)\in \mathbb{R}^2$. The \ac{LIP} dynamics along the $X$-axis is
%eq
\begin{equation}\label{eq:LIP_continuous}
    \frac{d}{dt}\begin{bmatrix} x\\ \dot{x} \end{bmatrix}= \begin{bmatrix} 0 & 1\\ \omega_o^2 &0 \end{bmatrix}\begin{bmatrix} x\\ \dot{x} \end{bmatrix}+\begin{bmatrix} \phantom{-}0\\ -\omega_o^2 \end{bmatrix}u^x
\end{equation}
%eq
where $\omega_o=\sqrt{g/H}$ and $g$ is the gravitational acceleration. By replacing $x$ with $y$ and $u^x$ with $u^y$ in \eqref{eq:LIP_continuous}, the dynamics along the $Y$-axis is derived.

Diagonalizing the state matrix in \eqref{eq:LIP_continuous} results in the following alternative choice of states
%eq
\begin{equation}\label{eq:DCM-CCM}
    \xi^x = x+\dot{x}/\omega_o ~~~\text{and}~~~  
    \eta^x = x-\dot{x}/\omega_o
    %x^{c} x^{ic}
\end{equation}
%eq
and integrating the resulting state-space representation gives 
%eq
\begin{align}
    \label{eq:DCM-evolution-x}
    \xi^{x}(t) &= e^{\omega_o t} \left( \xi^{x}(0)-u^x \right) + u^x \\
    \label{eq:CCM-evolution-x}
    \eta^{x}(t) &= e^{-\omega_o t} \left( \eta^{x}(0)-u^x \right) + u^x
    \enspace.
\end{align}
%eq
The unstable \eqref{eq:DCM-evolution-x} and stable \eqref{eq:CCM-evolution-x} solutions correspond, respectively, to the Divergent (\ac{DCM}) and Convergent (\ac{CCM}) Components of Motion; see~\cite{2015EnglsbergerThree} and Fig.~\ref{fig:all-figures}(b). Replacing the exponent ``$x$'' with ``$y$'' in \eqref{eq:DCM-evolution-x}-\eqref{eq:CCM-evolution-x} gives the analogous expressions in the $Y$ direction. Following~\cite{Righetti2020TRO}, we define $\xi-u$ and $\eta-u$ as the \ac{DCM} and \ac{CCM} offsets, respectively.

Assuming each step has duration $T$ and that the interval spanned by the $k$-th step is $[kT,(k+1)T)$, the \ac{LIP} state and input at the beginning of the $k$-th step are given by $\mathrm{x}_k = \begin{bmatrix} x_k & \dot{x}_k & y_k & \dot{y}_k \end{bmatrix}^\top$ and $\mathrm{u}_k = \begin{bmatrix} u^x_k & u^y_k \end{bmatrix}^\top$ yielding the following step-to-step dynamics
%eq
\begin{equation} \nonumber
    \mathrm{x}_{k+1} = A \mathrm{x}_k + B \mathrm{u}_k
\end{equation}
%eq
where $A = \text{diag}\{\hat{A}, \hat{A}\}$, $B = \text{diag}\{\hat{B}, \hat{B}\}$ and $\hat{A}, \hat{B}$ are block diagonal matrices obtained as detailed in~\cite{Narkhede2022}. We also define $\mathrm{y} = C \mathrm{x} = \begin{bmatrix} x & y\end{bmatrix}^\top$. The discrete evolution of the system must adhere to the robot's kinematic constraints, which can be captured by the foot reachability set
%eq
\begin{align}
    \nonumber
    \mathcal{U}_k = & \left\{ (u^x_k,u^y_k) ~|~ \underline{l}^x \leq u^x_k-u^x_{k-1} \leq \overline{l}^x \right. \\
    &~~~~~~~~~\text{and}~ \left. \underline{l}^y \leq m (u^y_k-u^y_{k-1}) \leq \overline{l}^y \right\}
    \label{eq:kin_const}
\end{align}
%eq
for $k \geq 1$, and $m = 1$ for a right stance foot and $m = -1$ otherwise. The rectangular region $[\underline{l}^x, \overline{l}^x] \times [\underline{l}^y, \overline{l}^y]$ approximates the set of swing-leg placements that are kinematically feasible when the current stance foot is located at $\mathrm{u}_{k-1} = (u^x_{k-1}, u^y_{k-1})$; see Fig.~\ref{fig:all-figures}(a).

The simplicity of the \ac{LIP} dynamics allows an explicit computation of the $N$-step capture region\footnote{Terminology: A state is \emph{$N$-step capturable} if there exists an evolution from it that brings the system to a stop within $N$ steps, respecting dynamic and actuation constraints. The \emph{$N$-step viable-capture basin} is the subset of states that are $N$-step capturable. An \emph{$N$-step capture point} is a ground location where a contact leads to an $(N-1)$-step capturable state; the set of all such points defines the \emph{$N$-step capture region}, $\mathcal{C}_N$; see~\cite{Koolen2011}.} in the form 
%eq
\begin{equation}\label{eq:N-step-capture-region}
    \mathcal{C}_N = \mathcal{C}^x_N \times \mathcal{C}^y_N
\end{equation}
%eq
for $N \in \{0,1,...\} \cup \{ \infty \}$. Indeed, Digit's symmetric foot placement in the $X$-direction where $\overline{l}^x=-\underline{l}^x = l^x$ results in symmetric $N$-step capture regions
%eq
\begin{equation}\label{eq:N-step-capture-region-x}
    \mathcal{C}^x_N = \{r \in \mathbb{R} ~|~ \underline{d}^x_N \leq r \leq \overline{d}^x_N \} 
\end{equation}
%eq
with bounds $\overline{d}^x_N = -\underline{d}^x_N = d^x_N$ computed recursively by~\cite{Koolen2011}
%eq
\begin{equation}\label{eq:capture-bounds-x}
    d^x_0=0 ~~\text{and}~~ d^x_N = e^{-\omega_o T} (l^x + d^x_{N-1})
    \enspace.
\end{equation}
%eq
The lateral asymmetry in Digit's foot placement makes the computation of the corresponding region in the $Y$-direction
%eq
\begin{equation}\label{eq:N-step-capture-region-y}
    \mathcal{C}^y_N = \{r \in \mathbb{R} ~|~ \underline{d}^y_N \leq r \leq \overline{d}^y_N \} 
\end{equation}
%eq
more involved; however, explicit expressions for the bounds $\underline{d}^y_N$ and $\overline{d}^y_N$ in \eqref{eq:N-step-capture-region-y} were recently derived in~\cite{Kumbhar2025ICRA}. These are omitted for brevity. Note also that the derivations in subsequent sections will focus on the $X$-direction; the corresponding $Y$-direction analysis is analogous, though more involved, and will be omitted for clarity.

%------------------------------------------------------------
\subsection{Capturability \& Restricted Spaces}
\label{subsec:capt-step-calc}
%------------------------------------------------------------

Consider a rectangular region $\mathcal{B} = [\underline{\beta}^x, \overline{\beta}^x] \times [\underline{\beta}^y, \overline{\beta}^y]$ within which the \ac{COM} must remain; see Fig.~\ref{fig:all-figures}(c). Enforcing this geometric constraint in a footstep planner ensures boundary collision avoidance. However, capturability is not guaranteed, motivating the development of a new constraint formulation. This can be achieved by computing a new rectangle $\mathcal{S}=[\underline{w}^x,\overline{w}^x]\times[\underline{w}^y,\overline{w}^y]$ inscribed in $\mathcal{B}$ so that restricting footsteps $\mathrm{u}=(u^x,u^y)$ in $\mathcal{S}$ guarantees both capturability and \ac{COM} containment in $\mathcal{B}$. For brevity, the construction of this rectangle is detailed below only for the $x$-direction; the $y$-direction is analogous.     

Without loss of generality, consider the first step in a sequence of steps; i.e., $k=0$. To ensure that the \ac{LIP}'s state is $N$-step capturable, we require that the \ac{DCM} offset satisfies $\xi^x_0 - u^x_0 \in \mathcal{C}^x_N$, which by \eqref{eq:N-step-capture-region-x} results in $\underline{d}^x_{N} \leq \xi^x_0 - u^x_0 \leq \overline{d}^x_{N}$. Multiplying this inequality with $e^{\omega_o t}$ and using \eqref{eq:DCM-evolution-x} gives the following bound for the \ac{DCM} evolution
%eq
\begin{equation} \nonumber
    e^{\omega_o t} \underline{d}^x_{N} \leq \xi^x(t) - u^x_0 \leq e^{\omega_o t} \overline{d}^x_{N}
\end{equation}
%eq
for $t \in [0,T)$. Next we turn our attention to the \ac{CCM} offset.
From~\cite{Kumbhar2025ICRA} we have that $-c^x \leq \eta^x_k - u^x_k \leq c^x$ for $k \geq 0$, where $c^x>0$ is explicitly computable. Multiplying this inequality for $k=0$ with $e^{-\omega_o t}$ and using \eqref{eq:CCM-evolution-x} gives
%eq
\begin{equation} \nonumber
    -e^{-\omega_o t} c^x \leq \eta^x(t) - u^x_0 \leq e^{-\omega_o t} c^x
\end{equation}
%eq
for $t \in [0,T)$. Combining these two inequalities and noting that $x(t) = \frac{1}{2}(\xi^x(t) + \eta^x(t))$ by \eqref{eq:DCM-CCM}, results in 
%eq
\begin{align}
    \nonumber
    \frac{e^{\omega_o t} \underline{d}^x_N-e^{-\omega_o t} c^x}{2} 
    \leq x(t) - u^x_0
    \leq \frac{e^{\omega_o t} \overline{d}^x_N+e^{-\omega_o t} c^x}{2}   
\end{align}
%eq
for $t \in [0,T)$, thus bounding the $X$-component of the \ac{COM} so that the first step is $N$-step capturable.

Now, to avoid collisions with the boundary of $\mathcal{B}$, the lower and upper bounds of the last inequality need to be such that $x(t) \in [\underline{\beta}^x, \overline{\beta}^x]$ for all $t \in [0,T)$; equivalently, 
%eq
\begin{align}
    \nonumber
    \underline{\beta}^x &\leq \inf_{t \in [0,T)} \frac{1}{2} (e^{\omega_o t} \underline{d}^x_N - e^{-\omega_o t} c^x) + u^x_0
    \\
    \nonumber
    \overline{\beta}^x &\geq \sup_{t \in [0,T)} \frac{1}{2} (e^{\omega_o t} \overline{d}^x_N + e^{-\omega_o t} c^x) + u^x_0 \enspace.
\end{align}
%eq
These conditions specify the admissible interval for $u^x_0$, with lower and upper bounds $\underline{w}^x$ and $\overline{w}^x$, respectively; that is $u^x_0 \in [\underline{w}^x, \overline{w}^x]$. The procedure for computing $\underline{w}^y$ and $\overline{w}^y$ is analogous and is omitted for brevity. The end result is the inscribed rectangle $\mathcal{S}$ shown in Fig.~\ref{fig:all-figures}(c), within which the foot can be placed to guarantee $N$-step capturability while avoiding boundary violations. The parameter $N$ governs the conservativeness of $\mathcal{S}$, with $N \to \infty$ yielding the most restrictive set.

%-----------------------------------------------------------
\subsection{Recursively Feasible \ac{MPC} in Rectangular Regions}
\label{subsec:rec-feas-mpc-rect}
%-----------------------------------------------------------

With the inscribed rectangle $\mathcal{S}=[\underline{w}^x,\overline{w}^x]\times[\underline{w}^y,\overline{w}^y]$ computed as above, an \ac{MPC} scheme can be formulated for footstep planning that simultaneously ensures capturability and boundary avoidance while providing recursive feasibility guarantees. Let $L$ denote the prediction horizon\footnote{Notation: To avoid ambiguity, $N$ denotes the number of steps used to compute capture regions and $L$ denotes the \ac{MPC} prediction horizon.} of the \ac{MPC} and for $\ell = 1,..., L$ define the running cost as
%eq
\begin{equation}\nonumber
    J_\ell(\mathrm{x}_\ell,\mathrm{u}_\ell) = \| \mathrm{x}_\ell - \mathrm{x}^\text{des} \|^2 + \| C \mathrm{x}_\ell - \mathrm{u}_\ell \|^2
\end{equation}
%eq
where $\mathrm{x}^{\mathrm{des}}$ is the desired \ac{LIP} state and the second term penalizes excessively long steps. The terminal cost is 
%eq
\begin{equation}\nonumber
    J_L(\mathrm{x}_L) = \| \mathrm{x}_L - \mathrm{x}^\mathrm{des} \|^2 \enspace.
\end{equation}
%eq
In our \ac{MPC} formulation, we set $L=N$; this choice results in less conservative sets.

To incorporate capturability into the \ac{MPC}, we first compute the maximum number of steps the \ac{LIP} can take within $\mathcal{S}$ along the forward $x^+$, backward $x^-$, rightward $y^+$, and leftward $y^-$ directions. For conciseness, we focus on the forward direction. The allowable number of steps equals the maximum-length $l^x$ steps the robot can take from its current foot position $u^x$ before reaching the boundary $\overline{w}^x$ of $\mathcal{S}$, i.e., $N^{x+} = \lfloor |\overline{w}^x - u^x|/l^x \rfloor$. Similar equations can be written for the other directions, namely, $N^{x-}, N^{y+}, N^{y-}$. Grouping the allowable number of steps in all directions in $\nu = [ N^{x+}~ N^{x-}~ N^{y+}~N^{y-} ]$, we define the \emph{$\nu$-step ($\mathcal{S}$-restricted) capture region} as
%eq
\begin{equation}\label{eq:c-nu}
    \mathcal{C}_\nu = \mathcal{C}^x_\nu \times \mathcal{C}^y_\nu
\end{equation}
%eq
where
%eq
\begin{align}
    \nonumber
    \mathcal{C}^x_\nu &= \left\{ r^x \in \mathbb{R} ~|~ \underline{d}^x_{N^{x-}} \!\leq\! r \leq \overline{d}^x_{N^{x+}} \right\}
    \label{eq:C-x-nu}
    \\
    \nonumber
    \mathcal{C}^y_\nu &= \left\{ r^y \in \mathbb{R} ~|~ \underline{d}^y_{N^{y-}} \!\leq\! r \leq \overline{d}^y_{N^{y+}}\right.~ \text{for right stance}
    \\
    \nonumber%\label{eq:C-y-nu}
    &\hspace{1.5cm}\left. -\overline{d}^y_{N^{y-}} \!\leq\! r^y \leq -\underline{d}^y_{N^{y+}}~ \text{for left stance}
    \right\}.
\end{align}
%eq

Given $\mathcal{C}_\nu$, we introduce a constraint in the \ac{MPC} that restricts the \ac{DCM} offset to lie within $\mathcal{C}_\nu$ with $\nu$ updated at each step to ensure both capturable evolution and boundary avoidance. The resulting \ac{MPC} formulation is
\vspace{0.2cm}
\noindent\rule{\columnwidth}{0.5pt}
\begin{align}
    &\minimize_{\mathbf{X}, \mathbf{U}, \mathbf{N}}\quad \sum_{\ell=k}^{k+L-1} J_\ell(\mathrm{x}_\ell, u_\ell)+J_L(\mathrm{x}_{k+L})
    \nonumber
    \\
    &\text{subject to}\quad\quad
    \mathrm{x}_{\ell+1} = A \mathrm{x}_\ell + B \mathrm{u}_\ell,
    ~\mathrm{x}_k = \mathrm{x}_\mathrm{init}
    \nonumber
    \\
    &\hspace{2cm} \nu_{\ell+1} = \nu_\ell - \mathbf{1}_4,
    ~\nu_k = \nu_\mathrm{init} - \mathbf{1}_4
    \nonumber
    \\
    &\hspace{2cm} \mathrm{u}_\ell \in \mathcal{U}_\ell
    \nonumber
    \\
    &\hspace{2cm} \xi_\ell - \mathrm{u}_\ell \in \mathcal{C}_{\nu_{\ell}}
    \nonumber
    \vspace{-0.2cm}
\end{align}
\noindent\rule{\columnwidth}{0.5pt}
%eq
\noindent where $\mathbf{1}_4 = [1~1~1~1]$, $\mathcal{U}_\ell$ is the constraint set \eqref{eq:kin_const}, and $\mathcal{C}_{\nu_{\ell}}$ is the $\mathcal{S}$-restricted capture region \eqref{eq:c-nu}. 
Starting from an initial condition $\mathrm{x}_\text{init}$ and allowable step count $\nu_\mathrm{init}$, the \ac{MPC} computes sequences $\mathbf{X} = \{\mathrm{x}_{k+1}, \ldots, \mathrm{x}_{k+L} \}$, $\mathbf{U} = \{\mathrm{u}_k, \ldots , \mathrm{u}_{k+L-1} \}$ and $\mathbf{N}=\{\nu_{k+1}, \ldots, \nu_{k+L}\}$. Only the first control input $\mathrm{u}_k$ is applied, after which the \ac{MPC} is re-solved at the next step using the updated $\mathrm{x}_\text{init}$ and $\nu_\mathrm{init}$. This process continues until the goal is reached.
Finally, the remark below provides conditions for recursive feasibility.

\begin{remark}\label{rem:rec-feas}
    The \ac{MPC} formulated above is recursively feasible, provided that $\mathcal{S}$ is non-empty and that $\nu_\mathrm{init}$, computed from the initial state and footstep locations as described in Section~\ref{subsec:capt-step-calc}, defines a viable-capture basin containing the \ac{LIP}'s initial state; see~\cite[Section IV-C]{Kumbhar2025ICRA} for details. Note that these recursive-feasibility guarantees apply only to the \ac{LIP}-based planning layer.
\end{remark}
%rem

%=============================================================
\section{Capturable Footstep Planning amid Obstacles}
\label{sec:appl-planning}
%=============================================================

This section generalizes the results of the previous section to rectangular corridors in three steps: (i) generating a collection of sequentially intersecting axis-aligned rectangles, (ii) defining a switching policy between them, and (iii) formulating a sequential \ac{MPC} with recursive feasibility guarantees via capturability, that provides footstep sequences that avoid obstacles and guide the robot to the goal.

%-------------------------------------------------------------
\subsection{Safe Corridor Generation}
\label{subsec:sc-generation}
%-------------------------------------------------------------

It is assumed that the environment $\mathcal{W}$ within which the robot operates is known. To highlight the core concepts and streamline the presentation, we consider here obstacles $O_i, i=1, ..., I,$ that can be represented as axis-aligned bounding boxes\footnote{Axis-aligned bounding boxes provide a simple geometric representation of obstacles in indoor and outdoor environments~\cite{Gao2018ICRA} and are widely used~\cite{Okada2005ICRA, Fang2022IROS} because they enable fast collision checks through linear inequalities.}, noting that more general obstacle shapes can also be incorporated. Let $\mathcal{W}^{\rm obs}$ be the union of all obstacles in $\mathcal{W}$.

Given the initial location of the robot, the task is to find a sequence of footsteps that take the robot to the final (desired) location, while avoiding obstacles on the way; that is, at each step $k$, the robot's \ac{COM} evolves inside $\mathcal{W} \setminus \mathcal{W}^{\rm obs}$. To achieve this, we present a systematic procedure for efficiently constructing \ac{SC}s, in the form of sequences of collision-free, axis-aligned rectangles
%eq
\begin{equation*}
    \mathcal{B} = \{\mathrm{y} \in \mathbb{R}^2 ~|~ P \mathrm{y} \leq b,~ P \in \mathbb{R}^{l \times 2},~b \in \mathbb{R}^{l} \}
\end{equation*}
%eq
that mutually intersect and connect the initial location with the goal; here the shorthand notation $P \mathrm{y} \leq b$ denotes the system of inequalities $r^\top_j \mathrm{y} \leq b_j$, $j \in \{1,...,l\}$ with $r^\top_j$ being the $j$-th row of $P$ and $b_j$ the $j$-th element of $b$.

%algorithm
\begin{algorithm}[b!]
\caption{$\mathtt{Generate\_Rectangles}$ $(\mathrm{p}_{\rm seed}, \{O_1,...,O_I\})$}\label{alg:safe_corridor}
\begin{algorithmic}[1]
\State Input: $\mathrm{p}_{\rm seed}$, $\{O_1,...,O_I\}$
\State Output: $\mathcal{B}$
\State $\mathcal{O} \gets \{O_1,..., O_I\}$, $i \gets 1$
    \While{$O_i \in \mathcal{O}$}
        \State $((r_i)^{\top}, (b_i)) \gets \mathtt{find\_hyperplane}(O_i, \mathrm{p}_{\rm seed})$
        \State $\mathcal{B} \gets \mathtt{append}(\mathcal{B}, \{((r_i)^{\top}, (b_i))\})$
        \State $\mathcal{O} \gets \mathtt{obstacle\_removal}(\mathcal{O}, \{((r_i)^{\top}, (b_i))\})$
        %\State $\mathcal{O} \gets \mathtt{obstacle\_removal}(\mathcal{O}, ((r_i)^{\top}, b_i))$
        \If{$\mathcal{O}$ is empty}
            \State \textbf{break}
        \EndIf
        \State $i \gets i+1$
    \EndWhile
\State \Return $\mathcal{B}$
\end{algorithmic}
\end{algorithm}
%algorithm

\begin{figure*}[t!]
\vskip +5pt
\centering
\subfigure[]{\includegraphics[width=0.23\textwidth]{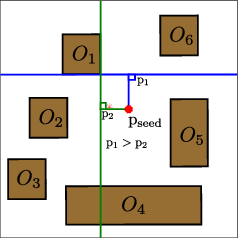}
\label{fig:Alg_1_Part2_a}}
\centering
\subfigure[]{\includegraphics[width=0.23\textwidth]{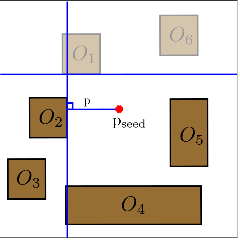}
\label{fig:Alg_1_Part2_b}}
\centering
\subfigure[]{\includegraphics[width=0.23\textwidth]{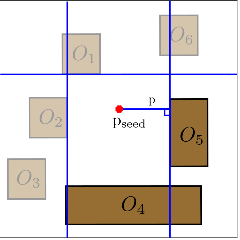}
\label{fig:Alg_1_Part2_c}}
\centering
\subfigure[]{\includegraphics[width=0.23\textwidth]{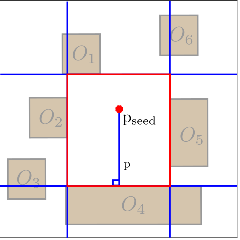}
\label{fig:Alg_1_Part2_d}}
\vskip -5pt
\caption{Rectangle generation via Algorithm~\ref{alg:safe_corridor}. Starting from ${\rm p}_{\rm seed}$, the algorithm iteratively selects the separating hyperplane farthest from the seed for each obstacle, intersecting the resulting half-spaces to build an obstacle-free rectangle. Redundant obstacles (e.g., $O_3$, $O_6$) are discarded at each iteration. The final rectangle (red) encloses a local obstacle-free region around ${\rm p}_{\rm seed}$.}
\label{fig:Algorithm_1}
\vskip -20pt
\end{figure*}

The construction of a safe corridor begins with a sequence of points $\Pi = \{\mathrm{p}_1,...,\mathrm{p}_K\} \in \mathcal{W} \setminus \mathcal{W}^{\rm obs}$ connecting the start and goal locations. These can be obtained via any sampling-based planner (e.g., \ac{JPS}~\cite{Liu2017RAL} or \ac{RRT}~\cite{LaValle2006Planning}), subject to the additional requirement that each consecutive line segment is obstacle-free. Given $\Pi$, we implement Algorithm~\ref{alg:safe_corridor}, which takes the obstacles $\{O_1,.., O_I\}$ and a seed point ${\rm p}_{\rm seed} \in \Pi$ as inputs and returns an obstacle-free rectangle $\mathcal{B}$. Algorithm~\ref{alg:safe_corridor} calls the following functions:
\begin{itemize}
\item $\mathtt{find\_hyperplane}$: Given a seed point $\mathrm{p}_{\rm seed}$ from $\Pi$ and the $i$-th obstacle $O_i$, this function returns a separating hyperplane that not only distinguishes the obstacle from the $\mathrm{p}_{\rm seed}$ but also maximizes its distance from it.
\item $\mathtt{obstacle\_removal}$: Given a list of obstacles and a hyperplane, this function discards those lying opposite the seed by determining the side of the hyperplane on which each obstacle's corner points lie.
\end{itemize}
Algorithm~\ref{alg:safe_corridor} proceeds as follows; see Fig.~\ref{fig:Algorithm_1} for a graphical illustration. At each iteration, the algorithm uses the seed point $\mathrm{p}_{\text{seed}}$ and the current obstacle set $\mathcal{O}$. It selects an obstacle from $\mathcal{O}$ and, via $\mathtt{find\_hyperplane}$, determines the halfspace of that obstacle farthest from $\mathrm{p}_{\text{seed}}$, which is then appended to $\mathcal{B}$. Next, $\mathtt{obstacle\_removal}$ updates $\mathcal{O}$ by discarding redundant obstacles, thereby shrinking the obstacle set at each step. This process continues until the obstacle list is empty, yielding an obstacle-free rectangle $\mathcal{B}$.

%fig
%fig

To construct the safe corridor, the procedure begins by using $\mathtt{Generate\_Rectangles}$ to generate a rectangle $\mathcal{B}_0$ that contains the initial \ac{COM} location, using it as the seed point. If $\mathcal{B}_0$ also contains the goal, the process terminates. Otherwise, the algorithm scans the points in $\Pi$ to identify the first one lying outside $\mathcal{B}_0$, which is then used as the new seed to compute the next rectangle $\mathcal{B}_{\mathrm{new}}$. The intersection $\mathcal{B}_0 \cap \mathcal{B}_{\mathrm{new}}$ is then checked: if the two rectangles intersect, the algorithm sets $\mathcal{B}_1 = \mathcal{B}_{\mathrm{new}}$; if not, the procedure in~\cite{Narkhede2022} is applied to connect $\mathcal{B}_0$ and $\mathcal{B}_{\mathrm{new}}$ through a sequence of mutually intersecting rectangles, computing the Chebyshev centers of their intersections and using them as seed points. By repeating this process, the algorithm yields a safe corridor comprising a sequence of mutually intersecting rectangles $\{\mathcal{B}_0, \mathcal{B}_1, \ldots, \mathcal{B}_{M-1}\}$, with $\mathcal{B}_0$ containing the initial location, $\mathcal{B}_{M-1}$ containing the goal location and $\mathcal{B}_{i} \cap \mathcal{B}_{i+1} \neq \emptyset$ for $i=0,..., M-2$.

%------------------------------------------------------------
\subsection{Extending Capturability to Safe Corridors}
\label{subsec:switching-policy}
%------------------------------------------------------------

The safe corridor $\{\mathcal{B}_0, \mathcal{B}_1, \ldots, \mathcal{B}_{M-1}\}$ constructed above is purely geometric: Constraining the robot's \ac{COM} to remain within it merely ensures collision avoidance. Our goal here is to incorporate capturability into locomotion across the entire corridor, which will enable recursive feasibility to extend from one rectangle to the full sequence. The proposed procedure consists of two stages, outlined below. First, new rectangles $\mathcal{S}_i$ inscribed in $\mathcal{B}_i$ are computed using the procedure in Section~\ref{subsec:capt-step-calc}. Constraining footsteps within ${\mathcal{S}}_i$ ensures both $N$-step capturability and that the \ac{COM} remains inside the original collision-free set $\mathcal{B}_i$.

Second, we extend capturability across the entire corridor to ensure recursive feasibility for the sequence of \ac{MPC} schemes that will be defined along it; see Section~\ref{subsec:seq-mpc-planner}  below. In prior safe-corridor approaches, transitions are purely geometric: Once the \ac{COM} entered the intersection of two consecutive safe regions, the feasibility region switches to the next. Such policies, however, neglect dynamics, and capturability is not guaranteed, even if it holds within each rectangle individually. To extend capturability across consecutive rectangles, a \emph{capturability-aware} switching policy is required, ensuring that at the moment of switching, there exists an evolution that brings the robot to a stop without exceeding the \emph{new} rectangle bounds.

\begin{figure}[t!]
\vskip +5pt
\centering
\includegraphics[width=0.95\columnwidth]{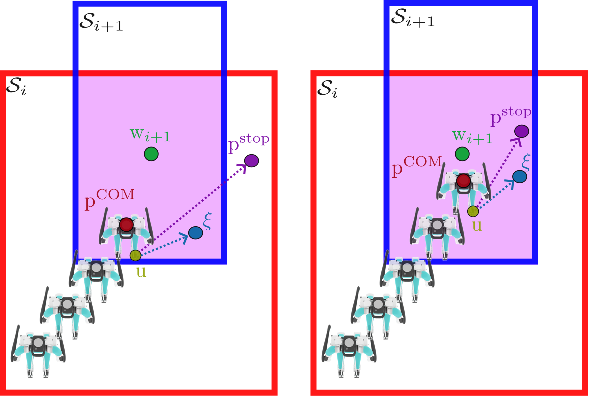}
\caption{Failed (left) vs successful (right) switching policy activation. Switching from ${\mathcal{S}}_{i}$ to ${\mathcal{S}}_{i+1}$ occurs once the virtual stopping point (purple), the \ac{COM} (red), and the current footstep (yellow) enter $\mathcal{S}_{i} \cap \mathcal{S}_{i+1}$. The current waypoint and \ac{DCM} are also shown in green and blue, respectively.}
\label{fig:poly-switch}
\vskip -10pt
\end{figure}

To formalize this notion, we define the \emph{virtual stopping point} $\mathrm{p}^\mathrm{stop} = (x^\mathrm{stop}, y^\mathrm{stop})$ as follows. Given that the \ac{LIP} is $N$-step capturable in $\mathcal{S}_i$, there exists a sequence of at most $N$ steps that, starting from the current state and footstep, brings the \ac{LIP} to a full stop. The virtual stopping point $\mathrm{p}^\mathrm{stop}$ is defined as the footstep location at which this complete stop occurs. Accordingly, ensuring capturability during the transition requires that---in addition to the \ac{COM} and current footstep---the virtual stopping point also lies within the intersection $\mathcal{S}_i \cap \mathcal{S}_{i+1}$ of the two rectangles; see Fig.~\ref{fig:poly-switch}. This way, capturability is preserved under switching, since a guaranteed evolution exists that brings the robot to rest without leaving the \emph{next} rectangle $\mathcal{S}_{i+1}$.

The virtual stopping point is directly determined by the \ac{DCM} offset and can be computed as follows. Focusing on the $X$-direction, to ensure $N$-step capturability, the $X$-component of the \ac{DCM} offset must be in $\mathcal{C}^x_N$ given by \eqref{eq:N-step-capture-region-x} with the relevant bounds provided by \eqref{eq:capture-bounds-x}. Specifically, suppose we have $\underline{d}^x_N \leq \xi^x - u^x \leq \overline{d}^x_N$. If $\xi^x - u^x \geq 0$, i.e., the \ac{DCM} is located in front of the current footstep, we find the smallest possible number $N'$ of maximum-length steps for which $\overline{d}^x_{N'}$ satisfies $\xi^x - u^x \leq \overline{d}^x_{N'}$. Then, the corresponding coordinate of the virtual stopping point is $x^\mathrm{stop} = u^x + N{'} \cdot l^x$ with $u^x$ the current footstep location and $l^x$ the maximum length of a step. The procedure is analogous for the case $\xi^x - u^x < 0$ with the difference that $N'$ is the smallest number of steps for which $\underline{d}^x_{N'} \leq \xi^x - u^x$ and $x^\mathrm{stop} = u^x - N' \cdot l^x$. The same procedure is applied in the $Y$-direction to determine the virtual stopping point coordinate $y^\mathrm{stop}$, although the lateral case is more complex due to the asymmetry in foot placement.

\begin{remark} \label{rem:stop-point}
The virtual stopping point $\mathrm{p}^{\mathrm{stop}}$ certifies the existence of a safe stopping configuration within $\mathcal{S}_{i} \cap \mathcal{S}_{i+1}$; it does \emph{not} necessarily require the robot to stop during transitions. In fact, as the robot approaches an intermediate goal, its state enters progressively lower-step capture regions, ultimately reaching the 1-step capture region. This ensures a transition to the next rectangle whenever sufficient free space is available for at least one maximum-length step, thereby preserving safety; otherwise, the robot stops at the intermediate goal, detecting that a collision with the boundary of the current rectangle $\mathcal{S}_{i}$ is imminent. Thus, the virtual stopping point functions as a feasibility certificate, effectively propagating recursive feasibility along the corridor.
\end{remark}
%rem

%fig
%fig

%------------------------------------------------------------
\subsection{A Sequential \ac{MPC} Planner with Recursive Feasibility}
\label{subsec:seq-mpc-planner}
%------------------------------------------------------------
 
We consider the ordered sequence of rectangles $\{ \mathcal{S}_0,\ldots, \mathcal{S}_{M-1} \}$ computed from the corresponding sequence of collision-free rectangles $\{ \mathcal{B}_0,\ldots, \mathcal{B}_{M-1}\}$ as in Section~\ref{subsec:switching-policy}. Assume that $\mathcal{S}_{i} \cap \mathcal{S}_{i+1} \neq \emptyset$ for all $i=0,\ldots,M-2$, and for each intersection $\mathcal{S}_{i} \cap \mathcal{S}_{i+1}$ compute the corresponding center of the Chebyshev disc $\mathrm{w}_{i+1}$, $i=0,...,M-2$, and let $\mathrm{w}_M$ denote the final destination. For generality, a desired velocity $\mathrm{v}_{i+1}$, $i=0,\dots,M-2$, is assigned to each waypoint, with zero velocity at the destination, i.e., $\mathrm{v}_M=0$. Thus, a sequence of triplets $(\mathcal{S}_i, \mathrm{w}_{i+1}, \mathrm{v}_{i+1})$, $i=0,\ldots,M-1$, is obtained and used to define a corresponding sequence \ac{MPC}$(i)$ of \ac{MPC} programs to plan footsteps with recursive feasibility guarantees that drive the robot to the destination. 

Specifically, \ac{MPC}$(i)$ is formulated as in Section~\ref{subsec:rec-feas-mpc-rect}, using the waypoint $(\mathrm{w}_{i+1}, \mathrm{v}_{i+1})$ to define the performance index. Given feedback from the \ac{COM} state $\mathrm{x}_\mathrm{init}$ and the stance foot location, the corresponding rectangle $\mathcal{S}_i$ is used to compute $\nu_\mathrm{init}$. \ac{MPC}$(i)$ is then solved to steer the \ac{LIP} toward the next waypoint, keeping the evolution within $\mathcal{S}_i$ and enforcing the kinematic constraint~\eqref{eq:kin_const}. Composing the \ac{MPC}s as 
%eq
\begin{equation}\nonumber
    \cdots \longrightarrow 
    \text{MPC}(i) \xrightarrow[]{\{\mathrm{p}^{\mathrm{COM}}_k,~\mathrm{u}_k,~\mathrm{p}^\mathrm{stop}_k\} \in \mathcal{S}_{i+1}} \text{MPC}(i+1)
    \longrightarrow \cdots
\end{equation}
%eq
for $i = 0,...,M-1$, with switching from \ac{MPC}$(i)$ to \ac{MPC}$(i+1)$ occurring when the \ac{COM} state, footstep location, and virtual stopping point all lie within $\mathcal{S}_{i+1}$ (cf. Section~\ref{subsec:switching-policy}), results in a sequence of suggested footstep locations that drive the \ac{LIP} to the final destination. 

Owing to the capturability-aware construction of the safe corridors, the sequential \ac{MPC} scheme described above offers recursive feasibility guarantees. To see this, assume that in the first call of \ac{MPC}$(0)$, $\mathcal{S}_0$ is non-empty and that the \ac{LIP}'s initial footstep location $\mathrm{u}_0$ and state $\mathrm{x}_0$ result in a $\nu_{\mathrm{init}}$ viable-capture basin that contains $\mathrm{x}_0$; $\nu_\mathrm{init}$ is computed based on $\mathcal{S}_0$ as in Section~\ref{subsec:rec-feas-mpc-rect}. Then, by Remark~\ref{rem:rec-feas}, \ac{MPC}$(0)$ is recursively feasible. Now, $\mathcal{S}_1$ is non-empty (by construction of the corridor, $\mathcal{S}_0 \cap \mathcal{S}_1 \neq \emptyset$) and the switching policy of Section~\ref{subsec:switching-policy} guarantees that, at the moment of switching, the \ac{LIP}'s state lies within the $\nu_\mathrm{init}$ viable-capture basin, with $\nu_\mathrm{init}$ computed by its current state, footstep location and $\mathcal{S}_1$. This follows from the fact that the conditions required to apply the policy ensure the existence of an evolution in $\mathcal{S}_1$ that can bring the \ac{LIP} to rest by eventually placing its footstep at the virtual stopping point. Hence, by Remark~\ref{rem:rec-feas}, \ac{MPC}$(1)$ is recursively feasible. The result follows by propagating these arguments forward until all rectangles in $\{\mathcal{S}_0,\dots,\mathcal{S}_{M-1}\}$ have been traversed.

\begin{remark} \label{rem:switched-velocity}
If the desired waypoint velocity is zero, the virtual stopping point is guaranteed to enter the intersection as the \ac{LIP} decelerates toward the waypoint. However, for nonzero velocities, the success of the switching policy depends on the interplay between the velocity and the intersection's size. For nonzero velocities, the success of the switching policy depends on the interplay between the velocity and the intersection's size. Gradually reducing the target velocity until the switching policy is activated practically eliminates this caveat.
\end{remark}

%=========================================================
\section{Results}
\label{sec:sim-res}
%=========================================================

We evaluate the \ac{LIP}-based sequential \ac{MPC}'s performance in random scenarios, focusing on success rates and computational times. We then evaluate the method in MuJoCo simulations with the bipedal robot \emph{Digit} navigating an office-like environment with door-opening constraints. In all results, the sequential \ac{MPC} is implemented in CasADi~\cite{Casadi} and solved using OSQP~\cite{osqp}. The suggested footsteps are realized via an Operational Space Control-based QP (OSC-QP). The objective of the OSC-QP is to track a set of weighted whole-body tasks, including Digit’s \ac{COM} following the \ac{LIP}, swing-foot trajectory tracking, torso orientation control, centroidal angular momentum minimization, and foot contact wrench optimization under friction cone constraints~\cite{Narkhede2022}. Note that  alternative low-level controllers can also be used, provided they sufficiently accurately track the planned \ac{LIP} dynamics and footsteps. The footstep \ac{MPC} runs at $15$\,Hz while the low-level OSC-QP operates at $400$\,Hz. The \ac{MPC} update rate of $15$\,Hz is selected to ensure approximately five \ac{MPC} iterations per Digit's step, assuming a step duration of $300\,\mathrm{ms}$, while also allowing sufficient time within each $1/400\,\mathrm{s}$ control cycle for the OSC-QP to execute the commanded footstep before the subsequent \ac{MPC} update. All computations were performed on an Intel PC with $\mathrm{i}7{\text -}9750\mathrm{H}$ processor ($2.60$ GHz) and $16\mathrm{GB}$ RAM.

%---------------------------------------------------------
\subsection{Numerical Experiments and Performance Evaluation}
\label{subsec:sim-compare}
%---------------------------------------------------------

To evaluate the efficiency of the proposed method, we generated $100$ random environments, each containing $10$ axis-aligned box obstacles covering $20$\% of a $50\,\mathrm{m} \times 50\,\mathrm{m}$ area, forming a sufficiently challenging set of environments. Across environments, the initial location is $(2\,\mathrm{m}, 2\,\mathrm{m})$, the destination is $(48\,\mathrm{m}, 48\,\mathrm{m})$, and both the initial and final velocities are zero. Waypoint velocities are set to Digit's allowable speed of $1\,{\rm m/s}$ for fast corridor traversal. For each environment, an \ac{AASC} is generated as in Section~\ref{subsec:sc-generation}, after which both the proposed capturability-aware sequential \ac{MPC} (Section~\ref{subsec:seq-mpc-planner}) and the \ac{MPC} scheme of~\cite{Narkhede2022} (hereafter, the prior method) are applied to the resulting \ac{SC}.

\begin{figure}
\vskip +5pt
\centering
\subfigure[]{\includegraphics[width=0.24\textwidth]{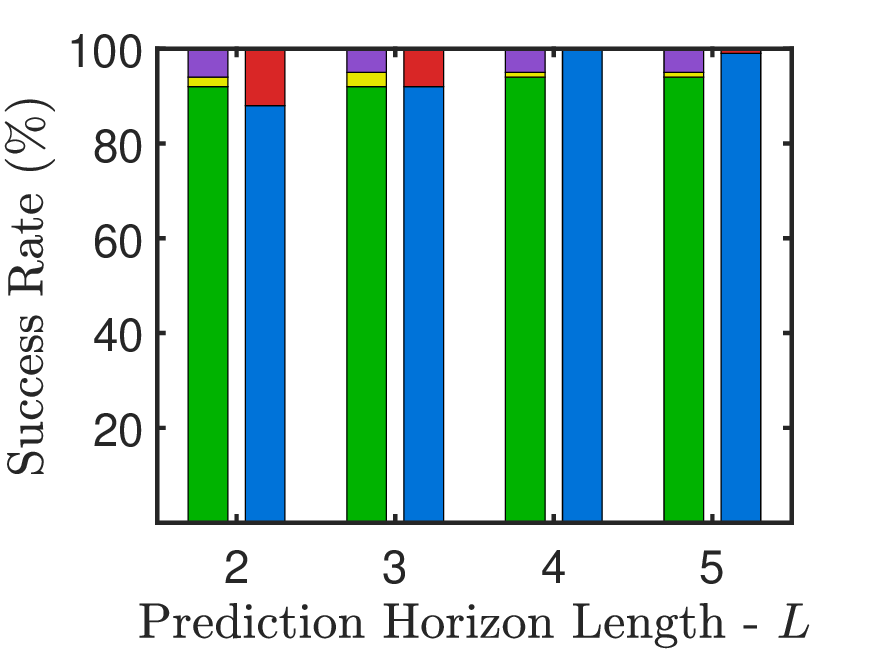} \label{fig:envs-res}}
\centering
\subfigure[]{\includegraphics[width=0.22\textwidth]{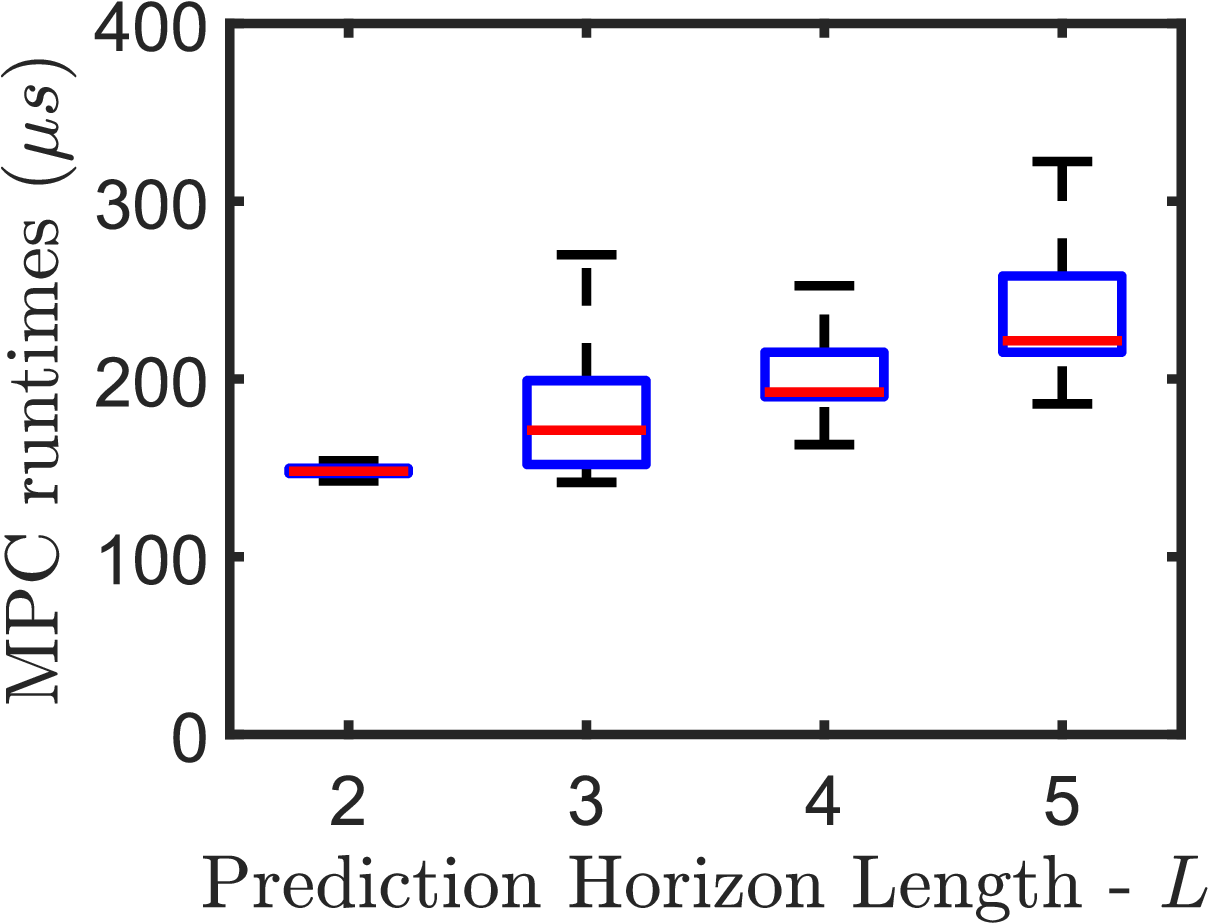} \label{fig:mpc-runtimes}}
\vskip -5pt
\caption{(a) \ac{MPC} success rates of the proposed and prior methods. Proposed method outcomes: success (green), in-place walking (yellow), and failure to compute a \ac{SC} (purple). Prior method outcomes: success (blue) and infeasibility. (b) \ac{MPC} runtimes for the proposed method.
}
\label{fig:polytopic_decomposition}
\vskip -15pt
\end{figure}

Fig.~\ref{fig:envs-res} directly compares the two methods. The proposed method maintains high success rates ($92\%-94\%$) across all prediction-horizon lengths, whereas the prior method achieves comparable performance only at longer horizons. Thus, the proposed scheme incurs lower computational cost, with runtimes on the order of $\mu$s that scale predictably with the horizon length (Fig.~\ref{fig:mpc-runtimes}). More significantly, Fig.~\ref{fig:envs-res} highlights the distinct modes through which each method fails to transfer the robot to the destination. The proposed method exhibits two such modes, both \emph{inherently safe}. The first arises when the capturability requirement results in inscribed rectangles that do not intersect, a condition that is detectable before execution. The second occurs in narrow or obstructed passages, causing the robot to resort to in-place walking when advancing may be dangerous. This is in stark contrast with the prior method, which may fail to suggest a footstep due to \ac{MPC} infeasibility. Hence, unlike the prior method, the proposed method always exhibits graceful, safety-preserving behaviors.

%----------------------------------------------------------
\subsection{Application to Office Spaces with Door Constraints}
\label{subsec:sim-mujoco}
%----------------------------------------------------------

\begin{figure*}[t!]
\vskip +5pt
\centering
\subfigure[]{\includegraphics[width=0.31\textwidth]{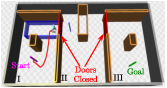}
\label{fig:Screenshot_1}}
\centering
\subfigure[]{\includegraphics[width=0.31\textwidth]{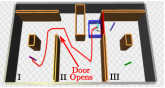}
\label{fig:Screenshot_2}}
\centering
\subfigure[]{\includegraphics[width=0.31\textwidth]{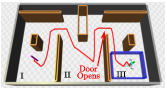}
\label{fig:Screenshot_3}}
\centering
\subfigure[]{\includegraphics[width=0.31\textwidth]{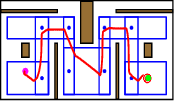}
\label{fig:map}}
\centering
\subfigure[]{\includegraphics[width=0.65\textwidth]{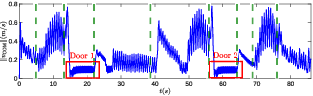}
\label{fig:v_com}}
\vspace{-0.15cm}
\caption{Digit navigating an office-like environment (rooms are labeled I, II and III) with two doors and three static obstacles. Start and goal points are in magenta and green; doors in red, obstacles in brown. The active \ac{SC} rectangle is shown in blue and the \ac{COM} trail in red. (b)-(c) Digit executing capturability-aware footsteps at closed doorways; upon door-opening, the planner resumes corridor execution. (d) Top view with capturability-aware \ac{SC}, start, and goal. (e) \ac{COM} speed: red boxes mark near-zero-velocity door-waiting phases; green dotted lines indicate entry into successive corridor intersections.}
\label{fig:mj-map}
\vskip -15pt
\end{figure*}

To show how the proposed method handles unexpected obstructions such as door constraints, consider the office-like environment shown in Fig.~\ref{fig:mj-map}, measuring $14{\rm m} \times 7{\rm m}$ with three rooms, two doors, and three static obstacles. Digit's task is to move from the first to the third room while avoiding obstacles and respecting door availability. The planner is \emph{not} informed of door-opening times in advance and assumes all doors are open at all times. We use a prediction horizon $L=5$ steps for the \ac{MPC}s. Figure~\ref{fig:mj-map}(a)–(c) show representative snapshots of Digit navigating within the capturability-aware safe corridor illustrated in Fig.~\ref{fig:mj-map}(d). Of particular interest is the planner's behavior with respect to door constraints highlighted in Figs.~\ref{fig:Screenshot_2}-\ref{fig:Screenshot_3}. When the next waypoint lies behind a \emph{closed} door, the \ac{MPC} guides the robot toward a capture state within the current rectangle. As shown in Fig.~\ref{fig:v_com}, the robot \emph{autonomously} slows down, switches to in-place walking and waits for the door to open. Once the door opens, the corridor extends through the doorway, becoming admissible and allowing the robot to continue toward the goal. Notably, the planner safely handles closed doors that temporarily render the destination unreachable, demonstrating robust navigation under exogenous, time-varying constraints while preserving capturability and recursive feasibility. Unlike the proposed method, the prior method lacks such guarantees and may therefore result in unsafe behaviors when encountering obstructions such as closed doors.

%fig
%fig

%fig
%fig

%=========================================================
\section{Conclusion}
%=========================================================

We presented a capturability-aware sequential \ac{MPC} footstep planner for humanoid locomotion in cluttered environments. The central contribution is a novel capturability-aware switching policy that propagates recursive feasibility across sequentially intersecting rectangles, extending capturability guarantees along the entire safe corridor. A key feature is that the robot autonomously decides not to proceed when a closed door blocks the path, reverting to in-place walking to avoid collision and waiting until the passage is clear. Moreover, success rates are largely independent of the prediction horizon, thus substantially reducing \ac{MPC} computational cost. The proposed framework lays a foundation for generating recursively feasible, collision-safe footstep plans for humanoids amid obstacles and under exogenous, time-varying constraints such as unknown door-opening times.

%=========================================================
\section*{Acknowledgment}
%=========================================================

\noindent
\begin{minipage}{0.14\linewidth}
\centering
\includegraphics[width=\linewidth]{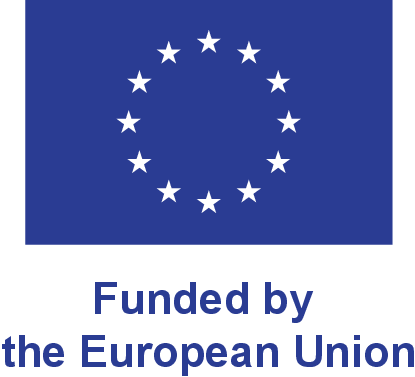}
\end{minipage}
\hfill
\begin{minipage}{0.82\linewidth}
\small
This project is funded by the European Union under Horizon Europe (grant No. 101136568 - HERON)
\end{minipage}

%=========================================================
\bibliographystyle{IEEEtran}
\bibliography{IEEEabrv, References}
%=========================================================

\end{document}